\pdfoutput=1
\documentclass[letterpaper, 10 pt, journal, twoside]{IEEEtran} 

\IEEEoverridecommandlockouts                              

\PassOptionsToPackage{dvipsnames}{xcolor}
\PassOptionsToPackage{table}{xcolor}

\newcommand{\subparagraph}{}
\usepackage[noindentafter]{titlesec}
\belowdisplayskip \abovedisplayskip

\usepackage{letltxmacro}

\LetLtxMacro{\oldsection}{\section}
\renewcommand{\section}[1]{
    \vspace{-0.11in}
    \oldsection{#1}
    \vspace{-0.10in}
}

\LetLtxMacro{\oldsubsection}{\subsection}
\renewcommand{\subsection}[1]{
    \vspace{-0.09in}
    \oldsubsection{#1}
    \vspace{-0.08in}
}

\LetLtxMacro{\oldsubsubsection}{\subsubsection}
\renewcommand{\subsubsection}[1]{
    \vspace{-0.06in}
    \oldsubsubsection{#1}
    \vspace{-0.05in}
}

\usepackage{times}
\usepackage{xspace}

\usepackage[pdftex]{epsfig}
\usepackage[pdftex]{graphicx}
\usepackage{wrapfig}

\usepackage{cite}

\usepackage{algorithm}
\usepackage{algorithmic}

\usepackage{paralist}

\usepackage{array}
\usepackage{multirow}
\usepackage[pdftex]{colortbl}
\usepackage[pdftex]{rotating}
\usepackage{makecell}
\usepackage{booktabs}
\usepackage{tabularx}

\usepackage{amsmath, amssymb}
\usepackage{textcomp}
\usepackage[pdftex]{stmaryrd}
\usepackage{upgreek}
\usepackage{bm}
\usepackage{cases}
\usepackage[pdftex]{mathtools}
\usepackage{arydshln} 
\usepackage{nicefrac}

\usepackage{url}
\usepackage[T1]{fontenc}
\usepackage[normalem]{ulem}
\usepackage{comment}
\usepackage[pdftex]{color}
\usepackage{balance}
\usepackage{framed}
\usepackage{epigraph}
\usepackage{soul}

\usepackage[pdftex]{adjustbox}%
\usepackage{textcomp}%
\usepackage{standalone}%
\usepackage{tikz}%
\usepackage{media9}
\usepackage{inconsolata}
\usepackage{fancyhdr}

\usepackage[hang,flushmargin]{footmisc}

\usepackage{macros/mysymbols}

\newcommand{\xhdr}[1]{\vspace{2pt}\noindent\textbf{#1}}

\newcommand{\sml}[1]{\textcolor{RubineRed}{#1}}%
\newcommand{\hpb}{HaPy\xspace}%
\newcommand{\hpbfull}{Habitat-PyRobot Bridge\xspace}%
\newcommand{\locobot}{LoCoBot\xspace}%
\newcommand{\srccSucc}{SRCC$_{\mbox{\tiny Succ}}$\xspace}%
\newcommand{\srccSPL}{SRCC$_{\mbox{\tiny SPL}}$\xspace}%

\newcommand{\tableTitle}[1]{\normalsize{#1}}%
\newlength{\figwidth}%

\newcolumntype{Y}{>{\centering\arraybackslash}X}
\newcolumntype{P}[1]{\begin{center}>{\arraybackslash}p{#1}\end{center}}

\usepackage[pagebackref=true,breaklinks=true,colorlinks,bookmarks=false]{hyperref}

\usepackage{cleveref}
\usepackage[font=small,labelfont=bf]{subcaption}
\usepackage{footmisc}
    
\begin{document}
\bstctlcite{IEEEexample:BSTcontrol}


\newcommand{\ourtitle}{
Sim2Real Predictivity: Does Evaluation in\\ Simulation Predict Real-World Performance?
\xspace}


\title{\ourtitle}

\author{Abhishek Kadian$^{1}$*, Joanne Truong$^{2}$*, Aaron Gokaslan$^{1}$, Alexander Clegg$^{1}$,
Erik Wijmans$^{1,2}$ \\
Stefan Lee$^{3}$,
Manolis Savva$^{1,4}$, Sonia Chernova$^{1,2}$
Dhruv Batra$^{1,2}$

\thanks{$^{1}$Albert Author is with Faculty of Electrical Engineering, Mathematics and Computer Science,
        University of Twente, 7500 AE Enschede, The Netherlands
        {\tt\small albert.author@papercept.net}}%
\thanks{$^{2}$Bernard D. Researcheris with the Department of Electrical Engineering, Wright State University,
        Dayton, OH 45435, USA
        {\tt\small b.d.researcher@ieee.org}}%
}


\twocolumn[{%
  \renewcommand\twocolumn[1][]{#1}%
  \maketitle
  \vspace{-0.5cm}
  \includegraphics[width=0.37\linewidth, trim=0cm -4cm 0cm 0cm]{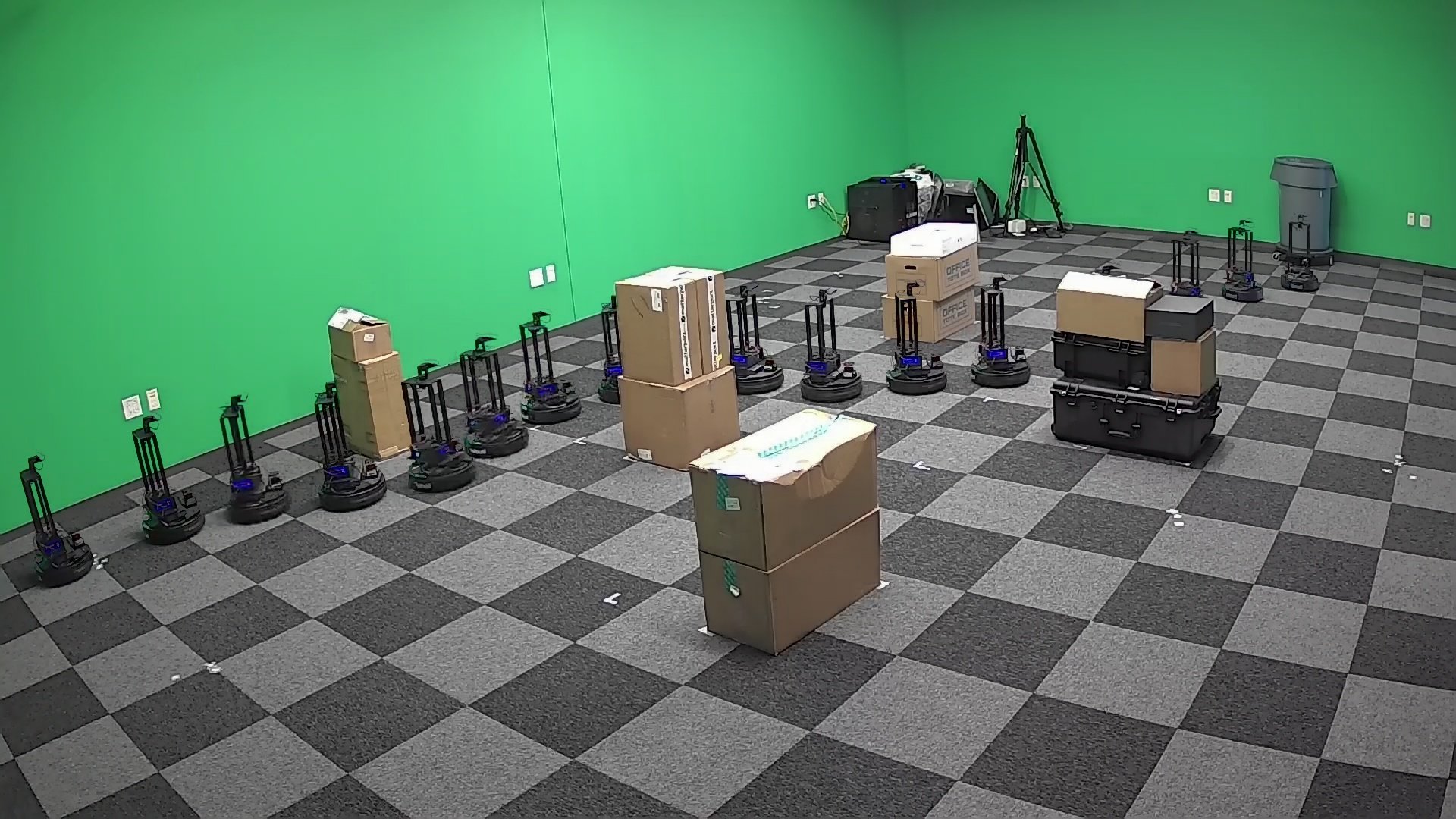}
  \includegraphics[width=0.37\linewidth, trim=0cm -4cm 0cm 0cm]{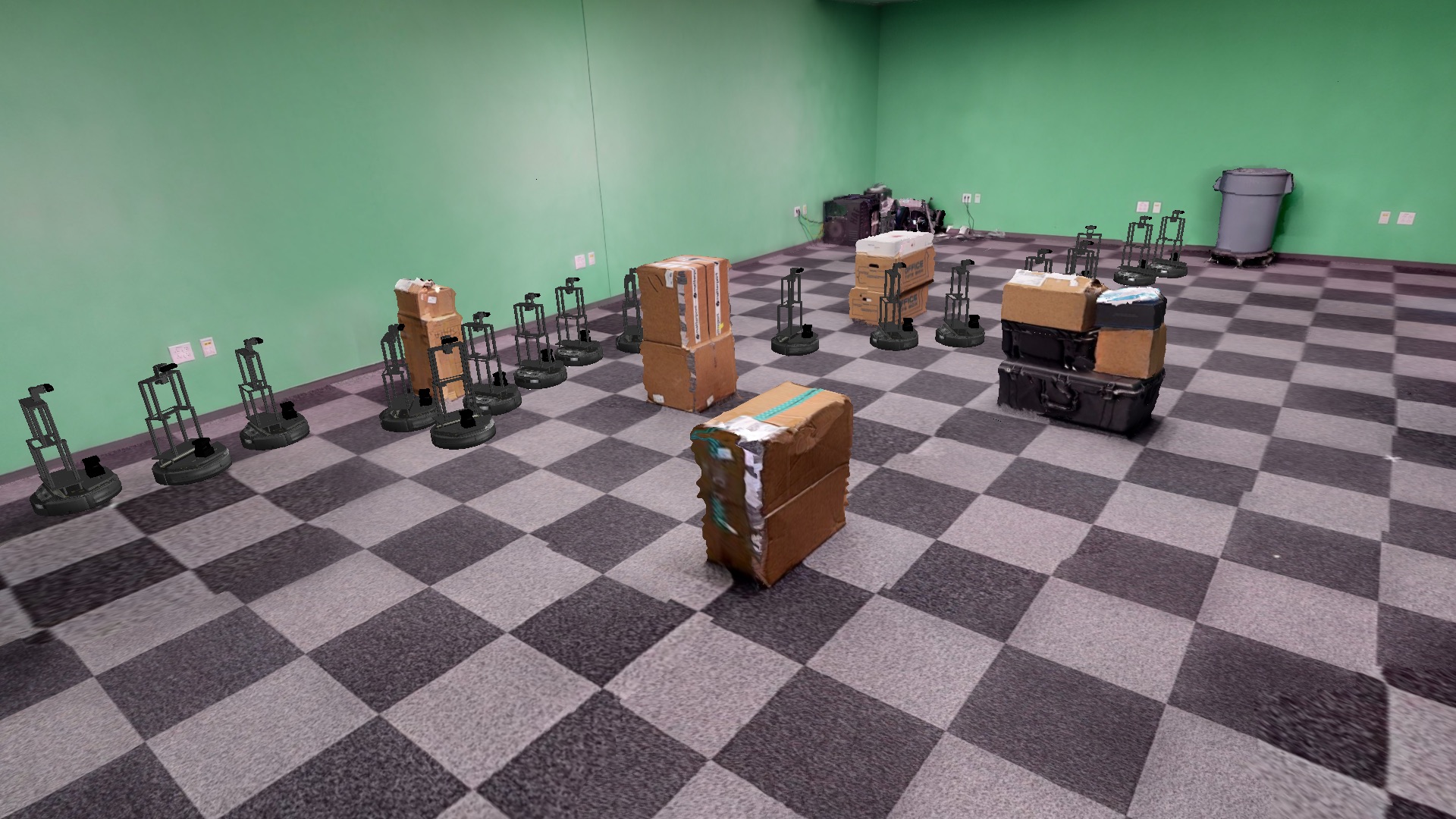}
  \includegraphics[width=0.25\linewidth, clip=true, trim=20pt 20pt 20pt 0pt]{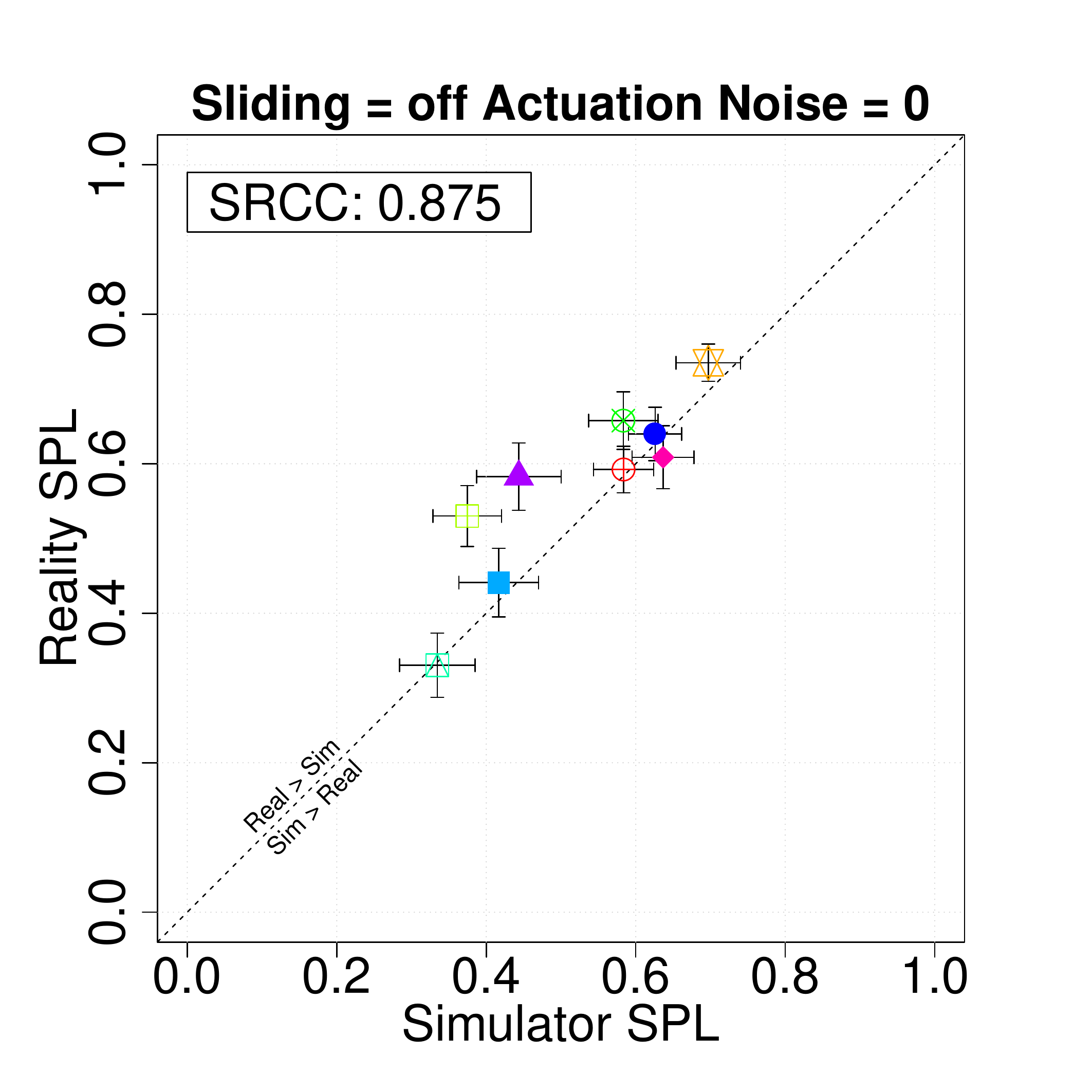} \\[-5pt] 
  \phantom{.} \hspace{0.7in} (a) Reality. \hspace{1.8in} (b) Simulation. \hspace{1in} (c) Sim-vs-Real Correlation. \\[-5pt]
  \label{fig:overview}
  \captionof{figure}{We measure the correlation between visual navigation performance in simulation and in reality by virtualizing reality and executing parallel experiments.
  (a): Navigation trajectory in a real space with obstacles.
  (b): virtualized replica in simulation. 
  (c): we propose the Sim2Real Correlation Coefficient (SRCC) as a measure of simulation predictivity.
  By optimizing for SRCC, we arrive at simulation settings that are highly predictive of real-world performance.
  \vspace{0.5cm}
  }
}]

\let\svthefootnote\thefootnote
\let\thefootnote\relax\footnote{
Manuscript received: February 24, 2020; Revised: May 23, 2020; Accepted: June 30, 2020.
\\
This paper was recommended for publication by Editor Eric Marchand upon evaluation of the Associate Editor and Reviewers’ comments.
\\
The Georgia Tech effort was supported in part by NSF, AFRL, DARPA, ONR YIPs, ARO PECASE, Amazon.
\\
\textsuperscript{1}AK, AG, AC, EW, MS, SC and DB are with Facebook AI Research {\tt\small\{akadian, agokaslan, alexclegg, erikwijmans, 
msavva, schernova, dbatra\}@fb.com}
\\
\textsuperscript{2}JT, EW, SC, DB are with Georgia Institute of Technology {\tt\small\{truong.j, etw, chernova, dbatra\}@gatech.edu}
\\
\textsuperscript{3}SL is with Oregon State University {\tt\small\{leestef\}@oregonstate.edu}
\\
\textsuperscript{4}MS is with Simon Fraser University {\tt\small\{msavva\}@sfu.ca}
\\
Digital Object Identifier (DOI): see top of this page.
}
\addtocounter{footnote}{-1}\let\thefootnote\svthefootnote\

\markboth{IEEE Robotics and Automation Letters. Preprint Version. Accepted June, 2020}
{Kadian \& Truong \MakeLowercase{\textit{et al.}}: Sim-2-Real Predictivity}   

 \vspace{-10pt}
\begin{abstract}
\noindent Does progress in simulation translate to progress on robots? If one method outperforms another in simulation, how likely is that trend to hold in reality on a robot? 
We examine this question for embodied PointGoal navigation -- developing engineering tools and 
a research paradigm for evaluating a simulator by its \emph{sim2real predictivity}.

First, we develop \hpbfull (\hpb), a library for seamless execution of  
identical code on simulated agents and robots -- transferring simulation-trained agents to a LoCoBot platform with a one-line code change.
Second, we investigate the sim2real predictivity of Habitat-Sim~\cite{habitat19iccv} for PointGoal navigation. 
We 3D-scan a physical lab space to create a \emph{virtualized replica}, 
and run parallel tests of 9 different models in reality and simulation. 
We present a new metric called Sim-vs-Real Correlation Coefficient (SRCC) to quantify predictivity. 

We find that SRCC for Habitat as used for the CVPR19 challenge is low ($0.18$ for the success metric), 
suggesting that performance differences in this simulator-based challenge do not persist after physical deployment. This gap is largely due to AI agents learning to exploit simulator imperfections -- abusing collision dynamics to `slide' along walls
, leading to shortcuts through otherwise non-navigable space. Naturally, such exploits do not work in the real world. 
Our experiments show that it is possible to \emph{tune} simulation parameters to improve sim2real predictivity (\eg improving \srccSucc from $\mathbf{0.18}$ to $\mathbf{0.844}$) -- increasing confidence that in-simulation comparisons will translate to deployed systems in reality.
\end{abstract} 

\begin{IEEEkeywords}
Visual-Based Navigation, Reinforcement Learning
\end{IEEEkeywords}

\section{Introduction}

\setlength{\epigraphwidth}{0.9\columnwidth}
\renewcommand{\epigraphflush}{center}
\epigraph{All simulations are wrong, but some are useful.}{\textit{A variant of a popular quote by George Box}
}
\vspace{-10pt}

\IEEEPARstart{T}{he} vision, language, and learning communities have recently witnessed a resurgence of interest
in studying integrative robot-inspired agents that perceive, navigate, and interact with their environment.
For a variety of reasons,
such work has commonly been carried out in simulation rather than in real-world environments.
Simulators can run orders of magnitude faster than real-time~\cite{habitat19iccv}, can be highly parallelized, and 
enable \emph{decades} of agent experience to be collected in days~\cite{ddppo}.
Moreover, evaluating agents in simulation is safer, cheaper, and enables easier 
benchmarking of scientific progress than running robots in the real-world.

Consequentially, these communities have rallied around simulation as a testbed -- developing several increasingly realistic indoor/outdoor navigation simulators~\cite{gupta2017cognitive,dosovitskiy2017carla,xia2018gibson,beattie2016deepmind,ai2thor,DBLP:SadeghiL17,habitat19iccv,savva2017minos},
designing a variety of tasks set in them~\cite{gupta2017cognitive,anderson2018vision,embodiedqa}, 
holding workshops about such platforms~\cite{simulation_workshop_eccv2018}, 
and even running challenges in these simulated worlds~\cite{habitat_challenge,carla_challenge,robothor_challenge}.
As a result, significant progress has been made in these settings. For example,
agents can reach point goals in novel home environments with near-perfect efficiency~\cite{ddppo},
control vehicles in complex, dynamic city environments~\cite{dosovitskiy2017carla},
follow natural-language instructions~\cite{anderson2018vision}, and answer questions~\cite{embodiedqa}.

However, no simulation is a perfect replica of reality, and AI systems are known to exploit imperfections and biases 
to achieve strong performance in simulation which may be unrepresentative of performance in reality. 
Notable examples include evolving tall creatures for locomotion that fall and somersault instead of learning 
active locomotion strategies \cite{lehman2018digitalcreativity} and OpenAI's hide-and-seek agents abusing their physics engine to `surf' on 
top of obstacles \cite{baker2019emergent}.

This raises several fundamental questions of deep interest to the scientific and engineering communities: 
\textbf{Do comparisons drawn from simulation translate to reality for robotic systems?} 
Concretely, if one method outperforms another in simulation, will it continue to do so when deployed on a robot?
Should we trust the outcomes of embodied AI challenges (\eg the AI Habitat Challenge at CVPR 2019)
that are performed entirely in simulation? These are questions not only of simulator \emph{fidelity}, but rather of \emph{predictivity}. 

In this work, we examine the above questions in the context of visual navigation -- focusing on measuring and optimizing the predictivity of a simulator. High predictivity enables researchers to use simulation for evaluation with confidence that the performance of different models will generalize to real robots. Given this focus, our efforts are orthogonal to techniques for sim2real transfer, including those based on adjusting simulator parameters. To answer these questions, we introduce engineering tools and a research paradigm for performing simulation-to-reality (sim2real) indoor navigation studies, revealing surprising findings about prior work.

First, we develop the \hpbfull (\hpb), a software library that enables seamless sim2robot transfer. \hpb is an interface between (1) Habitat~\cite{habitat19iccv}, a high-performance photorealistic 3D simulator, 
and (2) PyRobot~\cite{pyrobot2019}, a high-level python library for robotics research. 
Crucially, \hpb 
makes it trivial to \emph{execute identical code in simulation and reality}.
Sim2robot transfer with \hpb involves only a single line edit to the code (changing the \texttt{config.simulator} variable from \texttt{Habitat-Sim-v0} to \texttt{PyRobot-Locobot-v0}), 
essentially treating reality as just another simulator!  
This reduces code duplication, provides an intuitive high-level abstraction, and allows for rapid prototyping with modularity 
(training a large number of models in simulation and `tossing them over' for testing on the robot).
In fact, all experiments in this paper were conducted by a team of researchers physically separated by thousands of miles -- 
one set training and testing models in simulation, another conducting on-site tests with the robot, made trivial due to \hpb.
We will open-source \hpb so that everyone has this ability.


Second, we propose a general experimental paradigm for performing sim2real studies, which we call 
\emph{sim2real predictivity}. 
Our thesis is that simulators need not be a perfect replica of reality to be useful.
Specifically, we should primarily judge simulators not by their visual or physical realism, but by their 
\emph{sim2real predictivity} -- if method A outperforms B in simulation, how likely is the trend to hold in reality?
To answer this question, we propose the use of a quantity we call 
Sim2Real Correlation Coefficient (SRCC). 

We prepare a real lab space within which the robot must navigate while avoiding obstacles.
We then \emph{virtualize} this lab space (under different obstacle configurations) by 3D scanning the space 
and importing it in Habitat.
Armed with the power to perform parallel trials in reality and simulation, we test a suite of navigation models both in simulation and in the lab with a real robot. 
We then produce a scatter plot where every point is a navigation model, the x-axis is the performance in simulation, and the y-axis is performance in reality.
SRCC is shown in a box at the top.
If SRCC is high (close to 1), this is a `good' simulator setting in the sense that we can conduct scientific development and testing purely in simulation, with confidence that we are making `real' progress because the improvements in simulation will generalize to real robotic testbeds.
If SRCC is low (close to 0), this is a `poor' simulator, and we should have no confidence in results reported solely 
in simulation.
We apply this methodology in the context of PointGoal Navigation (PointNav)~\cite{anderson2018evaluation} with Habitat and the \locobot robot~\cite{locobot} as our simulation and reality platforms -- our experiments made easy with \hpb. 
These experiments reveal a number of surprising findings: 
\begin{compactitem}[\hspace{2pt}--]
\item We find that SRCC for Habitat as used for the CVPR19 challenge is $0.603$ for the Success
weighted by (normalized inverse) Path Length (SPL) metric and $0.18$ for agent success. When ranked by SPL, we observe $9$ relative ordering reversals from simulation to reality, suggesting that 
the results/winners may not be the same if the challenge were run on \locobot.

\item We find that large-scale RL trained models can learn to `cheat' by exploiting the way Habitat allows for `sliding' along walls on collision.
Essentially, the virtual robot is capable of cutting corners by sliding around obstacles, leading to unrealistic shortcuts through parts of non-navigable space and `better than optimal' paths.
Naturally, such exploits do not work in the real world where the robot stops on contact with walls.

\item We \emph{optimize} SRCC over Habitat design parameters 
and find that a few simple changes improve \srccSPL from $0.603$ to $0.875$ and \srccSucc from $0.18$ to $0.844$. The number of rank reversals nearly halves to $5$ ($13.8$\%).
Furthermore, we identify highly-performant agents in \emph{both} this new simulation and on \locobot in real environments.
\end{compactitem}

\noindent While our experiments are conducted on the PointNav task, we believe our software (\hpb), experimental paradigm (sim2real predictivity and SRCC), and take-away messages are 
useful to the broader community.
While we believe our controlled environment is complex enough to robustly estimate sim2real predictivity, we do not believe it should be used to measure navigation performance.
Our work is complementary to ongoing efforts to improve sim2real performance and our metric is independent of the  simulator implementation.

\section{Related Work}

\xhdr{Embodied AI tasks.}
Given the emergence of several 3D simulation platforms, it is not surprising that there has been a surge of research activity focusing on investigation of embodied AI tasks.
One early example leveraging simulation is the work of Zhu ~\etal~\cite{zhu2017target} on target-driven navigation using deep reinforcement learning in synthetic environments within AI2 THOR~\cite{ai2thor}.
Follow up work by Gupta \etal \cite{gupta2017cmp} demonstrated an end-to-end learned joint mapping and planning method evaluated in simulation using reconstructed interior spaces.
More recently, Gordon \etal \cite{gordon2019splitnet} showed that decoupling perception and policy learning modules can aid in generalization to unseen environments, as well as between different environment datasets.
Beyond these few examples, a breadth of recent work on embodied AI tasks demonstrates the acceleration that 3D simulators have brought to this research area.
In contrast, deployment on real robotic platforms for similar AI tasks still incurs significant resource overheads and is typically only feasible with large, well-equipped teams of researchers.
One of the most prominent examples is the DARPA Robotics Challenge (DRC)~\cite{krotkov2017darpa}.
Another example of real-world deployment is the work of Gandhi~\etal~\cite{gandhi2017learning} who trained a drone to fly in reality by locking it in a room.
Our goal is to characterize how well a model trained in simulation can generalize when deployed on a real robot.

\xhdr{Simulation-to-reality transfer.}
Due to the logistical limitations of physical experimentation, transfer of agents trained in simulation to real platforms is a topic of much interest.
There have been successful demonstrations of sim2real transfer in several domains.
The CAD2RL~\cite{DBLP:SadeghiL17} system of Sadeghi and Levine trained a collision avoidance policy entirely in simulation and deployed it on real aerial drones.
Similarly, Muller~\etal~\cite{muller2018driving} show that driving policies can be transferred from simulated cars to real remote-controlled cars by leveraging modularity and abstraction in the control policy. 
Tan~\etal~\cite{tan2018sim} train quadruped locomotion policies in simulation by leveraging domain randomization and demonstrate robustness when deployed to real robots.
Chebotar~\etal~\cite{simopt} improve the simulation using the difference between simulation and reality observations.
Lastly, Hwangbo~\etal~\cite{hwangbo2019learning} train legged robotic systems in simulation and transfer the learned policies to reality.
The goal of this work is to enable researchers to use simulation for evaluation with confidence that their results will generalize to real robots.
This brings to the forefront the key question: can we establish a correlation between performance in simulation and in reality?
We focus on this question in the domain of indoor visual navigation.

\section{Habitat-PyRobot Bridge: Simple Sim2Real}
\label{sec:hapy}

\noindent Deploying AI systems developed in simulation to physical robots presents significant financial, engineering, and logistical challenges -- especially for non-robotics researchers.
Approaching this directly requires researchers to maintain two parallel software stacks, one typically based on ROS~\cite{quigley2009ros} for the physical robot and another for simulation.
In addition to requiring significant duplication of effort, this model can also introduce inconsistencies between agent details and task specifications in simulation and reality.

To reduce this burden and enable our experiments, we introduce the Habitat-PyRobot Bridge (HaPy). 
As its name suggests, HaPy integrates the Habitat~\cite{habitat19iccv} platform with PyRobot APIs~\cite{pyrobot2019} -- enabling identical agent and evaluation code to be executed in simulation with Habitat and on a PyRobot-enabled physical robot.
Habitat is a platform for embodied AI research that aims to standardize the different layers of the embodied agent software stack, covering 1) datasets, 2) simulators, and 3) tasks. 
This enables researchers to cleanly define, study, and share embodied tasks, metrics, and agents.
For deploying simulation-trained agents to reality, we replace the simulator layer in this stack with `reality' while maintaining task specifications and agent interfaces.
Towards this end, we integrate Habitat with PyRobot \cite{pyrobot2019}, a recently released high-level API that implements simple interfaces to abstract lower-level control and perception for mobile platforms (LoCoBot), and manipulators (Sawyer), and offers a seamless interface for adding new robots. 
HaPy is able to benefit from the scalability and generalizability of both Habitat and PyRobot.
Running an agent developed in Habitat on the LoCoBot platform requires changing a single argument, \texttt{Habitat-Sim-v0}$\rightarrow$\texttt{PyRobot-Locobot-v0}.

\noindent At a high level, HaPy enables the following:\\[-15pt]
\begin{compactenum}[\hspace{3pt}--]
    \item A uniform observation space API across simulation and reality. Having a shared implementation ensures that observations from simulation and reality sensors go through the same transformations (e.g. resizing, normalization).
    \item A uniform action space API for agents across simulation and reality. Habitat and PyRobot differ in their agent action spaces. Our integration unifies the two action spaces and allows an agent model to remain agnostic between simulation and reality.
    \item Integration at the simulator layer in the embodied agent stack allows reuse of functionalities offered by task layers of the stack -- task definition, metrics, \etc stay the same across simulation and reality. 
    This also opens the potential for \emph{jointly} training in simulation and reality.
    \item Containerized deployment for running challenges where participants upload their code for seamless deployment to mobile robot platforms such as \locobot.
\end{compactenum}
\noindent We hope this contribution will allow the community to easily build agents in simulation and deploy them in the real world. 
\section{Visual Navigation in Simulation \& Reality}


\noindent Recall that our goal is to answer an ostensibly straightforward scientific question -- is performance in simulated environments predictive of real-world performance for visual navigation? Let us make this more concrete. 

First, note this question is about \emph{testing} in simulation vs reality.
It does not require us to take a stand on \emph{training} in simulation vs reality (or the need for training at all).
For a comparison between simulation-trained and non-learning-based navigation methods, we refer the reader to previous
studies~\cite{mishkin2019benchmarking, kojima2019learn, habitat19iccv}. 
We focus on test-time discrepancies 
between simulation and reality for learning-based methods. 

Second, even at test-time, many variables contribute to the sim2real gap. The real-world test environment may include objects or rooms which visually differ from simulation, 
or may present a differing task difficulty distribution 
(due to unmodeled physics or rendering in simulation).
To isolate these factors as much as possible, we propose a direct comparison -- evaluating agents in physical environments and in corresponding simulated replicas.
We construct a set of physical lab environment configurations for a robot to traverse and virtualize each by 3D scanning the space, thus controlling for semantic domain gap. 
We then evaluate agents in matching simulated and real configurations to characterize the sim-vs-real gap in visual navigation.

In this section, we first recap the PointNav task \cite{anderson2018evaluation} from the recent AI Habitat Challenge \cite{habitat_challenge}.
Then, we compare agents in simulation and robots in reality. 


\subsection{Task: PointGoal Navigation (PointNav)}

In this task, an agent is spawned in an unseen environment and asked to navigate to a goal location specified in relative coordinates.
We start from the agent specification and observation settings from \cite{habitat_challenge}.
Specifically, agents have access to 
an egocentric RGB (or RGBD) sensor and 
accurate localization and heading via a GPS+Compass sensor. 
The goal is specified using polar coordinates $(r,\theta)$, 
where $r$ is the Euclidean distance to the goal 
and $\theta$ is the azimuth to the goal. 
The action space for the agent is: \texttt{turn-left 30$^{\circ}$}\footnote{\label{angle} Originally 10$^{\circ}.$}, 
\texttt{turn-right 30$^{\circ}$}\footref{angle}, \texttt{forward 0.25m}, and \texttt{STOP}.
An episode is considered successful if the agent issues the \texttt{STOP} command within $0.2$ meters of the goal. 
Episodes lasting longer than $200$ steps or calling the \texttt{STOP} command >$0.2\text{m}$ from goal are declared unsuccessful.
We set the step threshold to $200$ (compared to $500$ in the challenge) because our testing lab is small and we found that episodes longer than $200$ actions are likely to fail.
We also limit collisions to $40$ to prevent damage to the robot.
We find that ${>}40$ collisions in an episode typically occur when the robot is stuck and likely to fail.

\vspace{-.1cm}
\subsection{Agent in Simulation}

\begin{figure}
  \centering%
  \resizebox{\columnwidth}{!}{
  \renewcommand{\tableTitle}[1]{\large{#1}}%
  \setlength{\figwidth}{0.3\columnwidth}%
  \setlength{\tabcolsep}{1.5pt}%
  \renewcommand{\arraystretch}{0.8}%
  \renewcommand\cellset{\renewcommand\arraystretch{0.8}%
  \setlength\extrarowheight{0pt}}%
  
  
  \hspace{-0.25cm}\begin{tabular}{c c c c c}
   &\tableTitle{Robot}&\tableTitle{RGB}&\tableTitle{Depth}&\tableTitle{Trajectories}\\
   \rotatebox[origin=c]{90}{\tableTitle{Reality}}&%
   \makecell{\includegraphics[width=\figwidth]{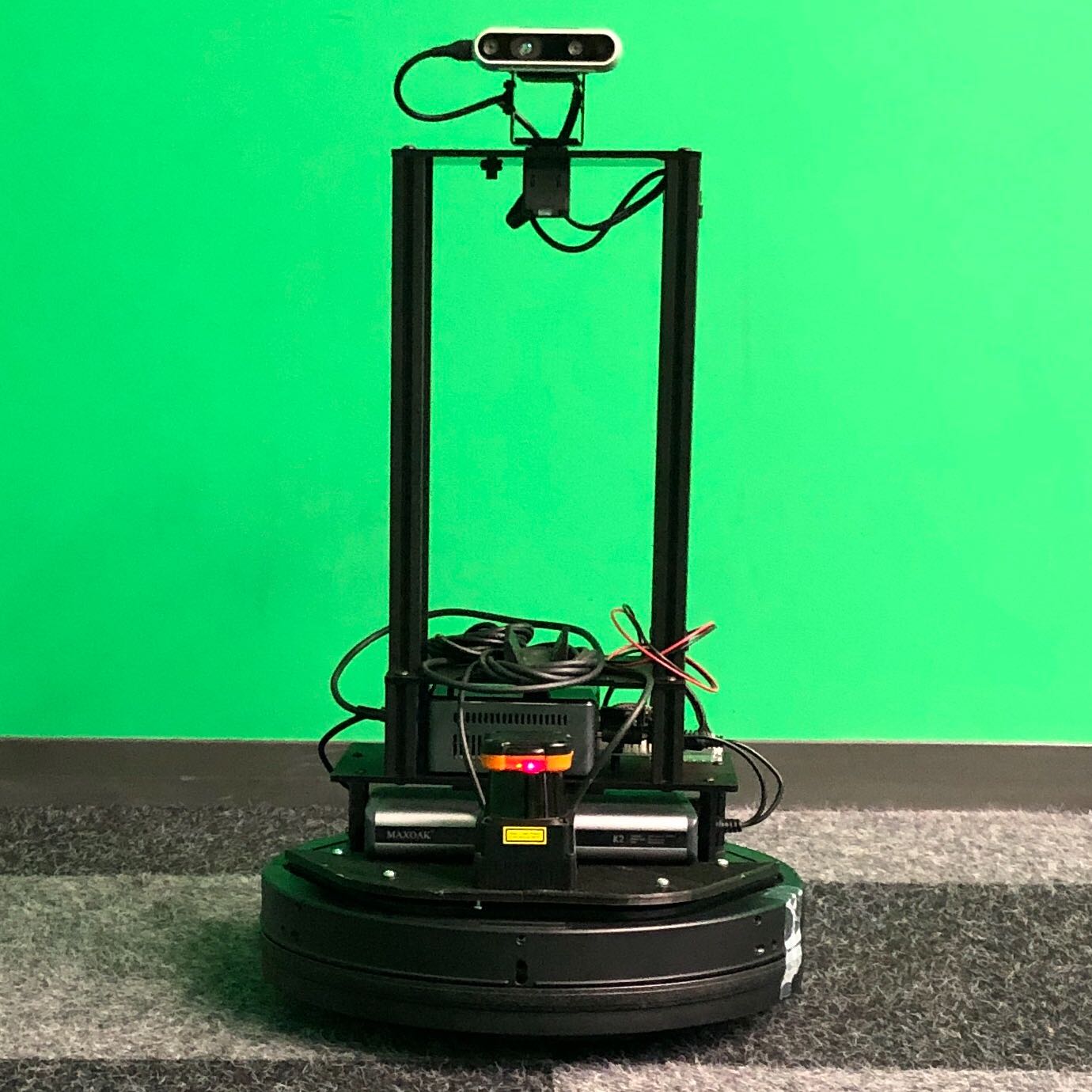}}&%
   \makecell{\includegraphics[width=\figwidth]{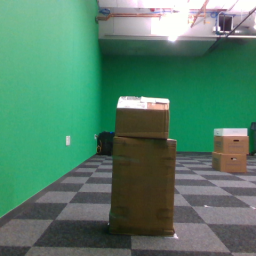}} &
   \makecell{\includegraphics[width=\figwidth]{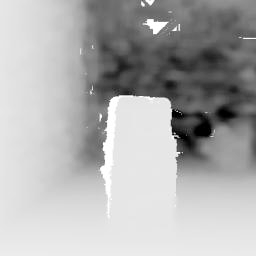}} &
   \makecell{\includegraphics[width=\figwidth]{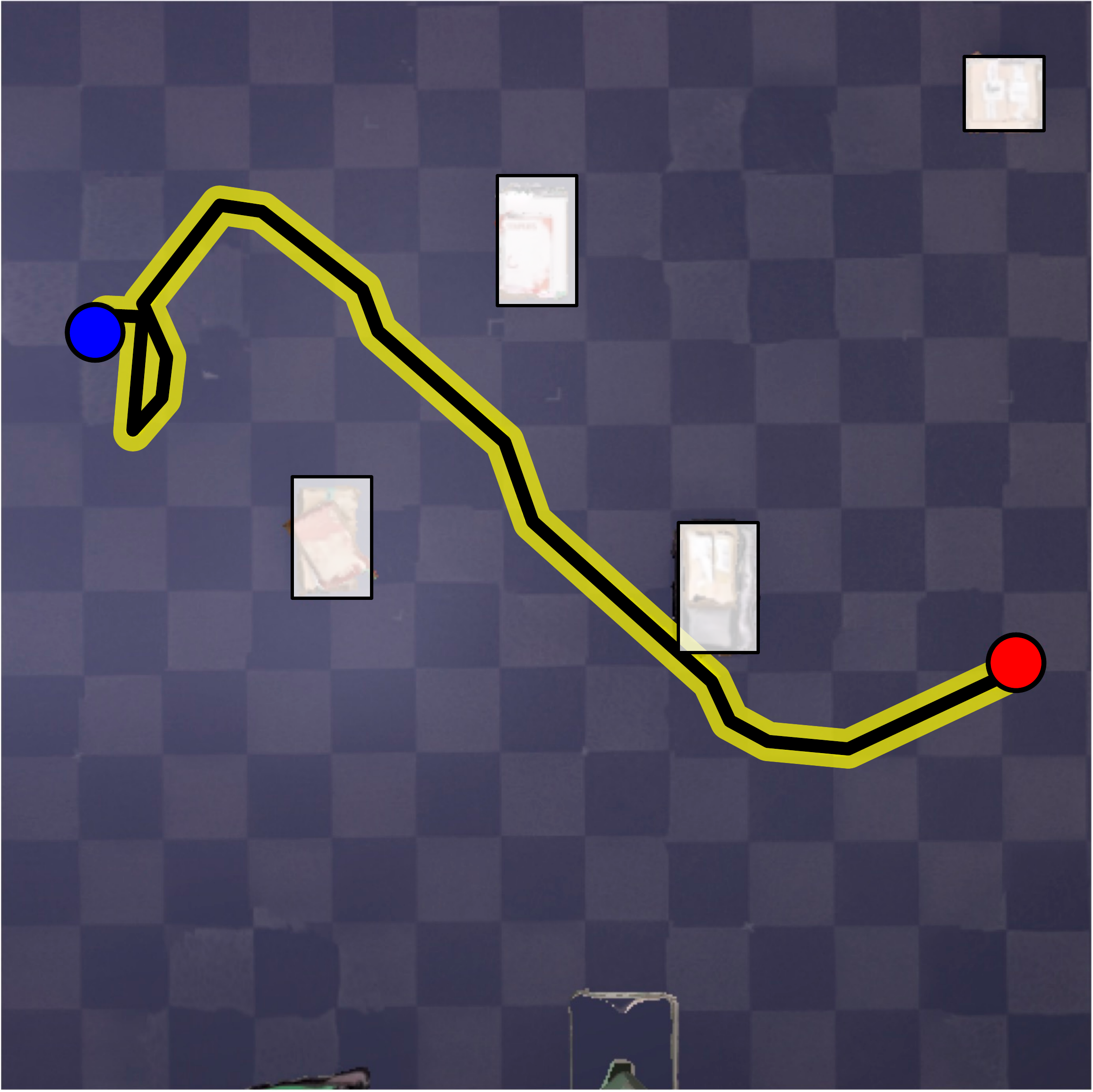}} \\
   \rotatebox[origin=c]{90}{\tableTitle{Simulation}}&%
    \makecell{\includegraphics[width=\figwidth]{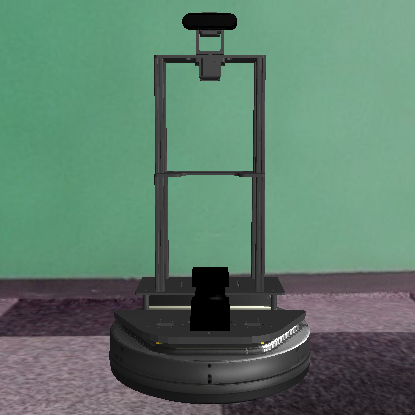}}&
   \makecell{\includegraphics[width=\figwidth]{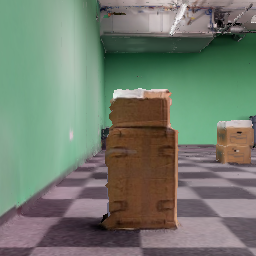}}&%
   \makecell{\includegraphics[width=\figwidth]{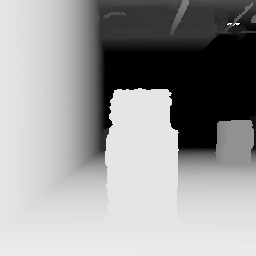}} &
   
   \makecell{\includegraphics[width=\figwidth]{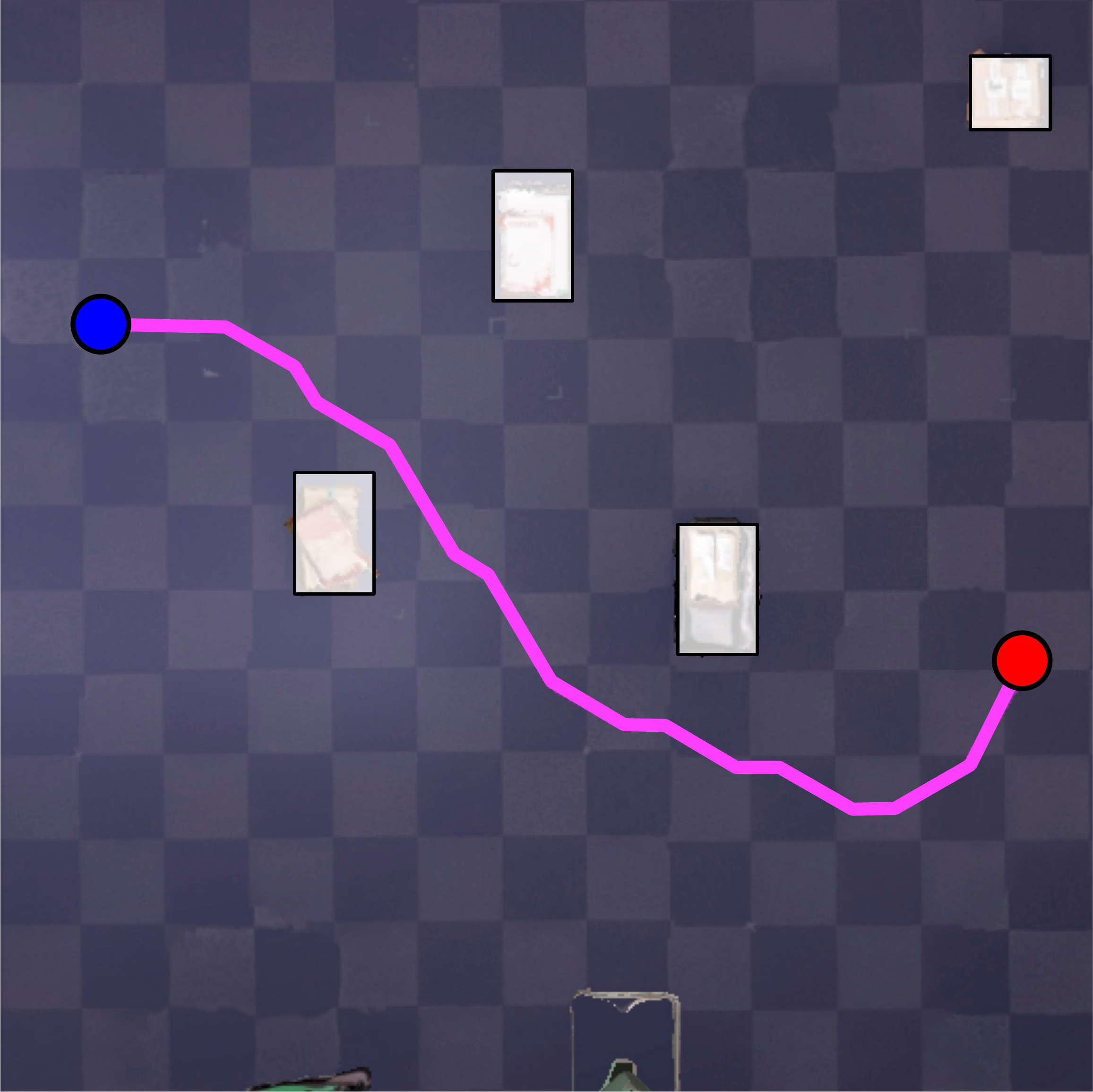}}\\
  \end{tabular}}
  \caption{Simulation vs.~reality.  Shading on the trajectory in reality represents uncertainty in the robot's location ($\pm$7cm).}
  \label{fig:simvreal}
  \end{figure}

\xhdr{Body.}
The experiments by Savva~\etal~\cite{habitat19iccv} 
and the Habitat Challenge 2019 \cite{habitat_challenge} model the agent as an idealized cylinder 
of radius $0.1$m and height $1.5m$. 
As shown in \figref{fig:simvreal}, 
we configure the agent to match the robot used in our experiments (\locobot) 
as closely as possible.
Specifically, we configure the simulated agent's base-radius and height to be $0.175\text{m}$ and $0.61\text{m}$ respectively to match \locobot dimensions. 

\xhdr{Sensors.}
We set the agent camera field of view to $45$ degrees to match images from the Intel D435 camera on \locobot.
We match the aspect ratio and resolution of the simulated sensor frames to real sensor frames from \locobot using square center cropping followed by image resizing to a height and width of 256 pixels.
To mimic the depth camera's limitations, 
we clip simulated depth sensing to $10\text{m}$.

\begin{figure}
    \centering
    \includegraphics[width=0.85\columnwidth]{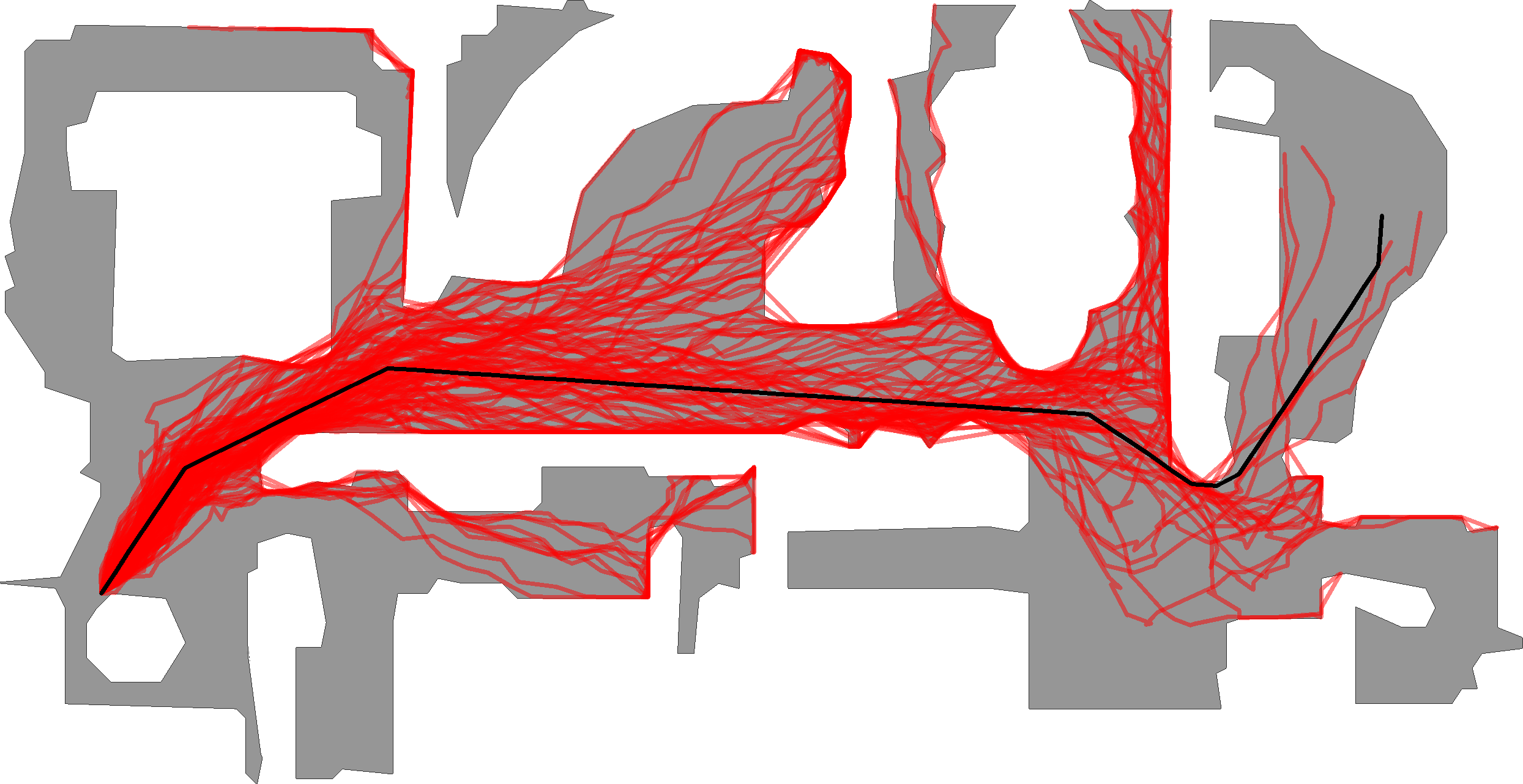}
    \caption{Effect of actuation noise. The black line is a trajectory from an action sequence with perfect actuation. In red are trajectories from this sequence with actuation noise.}
    \label{fig:noisy-actions}
\end{figure}

\xhdr{Actions.}
In \cite{habitat19iccv, habitat_challenge}, 
agent actions are deterministic -- \ie when the agent executes \texttt{turn-left 30$^{\circ}$}, it turns \emph{exactly} $30^\circ$, and \texttt{forward 0.25\text{m}} moves the agent \emph{exactly} $0.25\text{m}$ forward (modulo collisions).
However, no robot moves deterministically due to real-world actuation noise.  
To model the actions on \locobot, we leverage an actuation noise model derived from mocap-based benchmarking 
by the PyRobot authors~\cite{pyrobot2019}. 
Specifically, when the agent calls (say) \texttt{forward}, 
we sample from 
an action-specific 2D Gaussian distribution over relative displacements. \figref{fig:noisy-actions} shows  
trajectory rollouts sampled from this noise model. 
As shown, identical action sequences can lead to vastly different final locations. 

Finally, in contrast to \cite{habitat19iccv, habitat_challenge}, 
we increase the angles associated with  \texttt{turn-left} and \texttt{turn-right} actions from $10^{\circ}$ to $30^{\circ}$ degrees. 
The reason is a fundamental discrepancy between 
simulation and reality -- there is no `ideal' GPS+compass sensor in reality. Perfect localization in indoor environments is an open research problem. 
In our preliminary experiments, we found that 
localization noise was exacerbated by 
the `move, stop, turn, move' behavior of the robot, 
which is a result of a discrete action space (as opposed to continuous control via velocity or acceleration actions). We strike a balance between staying 
comparable to prior work (that uses discrete actions) 
and reducing localization noise by increasing the turn angles (which decreases 
the number of robot restarts). 
In the longer term, we believe the community should move towards continuous control to overcome this issue. 
All the modeling parameters are easily adaptable to different robots.

\label{sec:sim_agent_config}

\vspace{-.1cm}
\subsection{\locobot in Reality}
\label{sec:locobot}

\vspace{-.1cm}
\xhdr{Body.}
\locobot is designed to provide easy access to a robot with basic grasping, locomotion, and perception capabilities.
It is a modular robot based on a Kobuki YMR-K01-W1 mobile base with an extensible body. 

\vspace{-.1cm}
\xhdr{Sensors.}
\locobot is equipped with a Intel D435 RGB+depth camera. 
While \locobot possesses on-board IMUs and motor encoders
(which can provide the GPS+Compass sensor observations required by this task), the frequent stopping and starting from our discrete actions resulted in significant error accumulation.
To provide precise localization, we mounted a Hokuyo UTM-30LX LIDAR sensor in place of the robot's grasping arm (seen in \Cref{fig:simvreal}).
We run the LIDAR-based Hector SLAM~\cite{hector_slam} algorithm to provide the location+heading 
for the GPS+Compass sensor and for computing success and SPL 
of tests in the lab.

At this point, it is worth asking how accurate the 
LIDAR based localization is. 
To quantify localization error, we ran a total of $45$ tests across $3$ different room configurations in the lab, 
and manually measured the error with measuring tape. 
On average, we find errors of approximately $7\text{cm}$ with Hector SLAM, compared to $40\text{cm}$ 
obtained from wheel odometry and onboard IMU (combined 
with an Extended Kalman Filter implementation in ROS). 
Note that $7cm$ is significantly lower than the $0.2m = 20cm$ criteria used to define success in PointNav, 
providing us confidence that we can use LIDAR-based Hector SLAM to judge success in our real-world experiments.  
More importantly, we notice that the LIDAR approach allows the robot to reliably relocalize using its surroundings, and thus error does not accumulate over long trajectories, or with consecutive runs, which is important for running hundreds of real-world experiments.

\begin{figure}
    \centering
    \includegraphics[width=0.9\columnwidth]{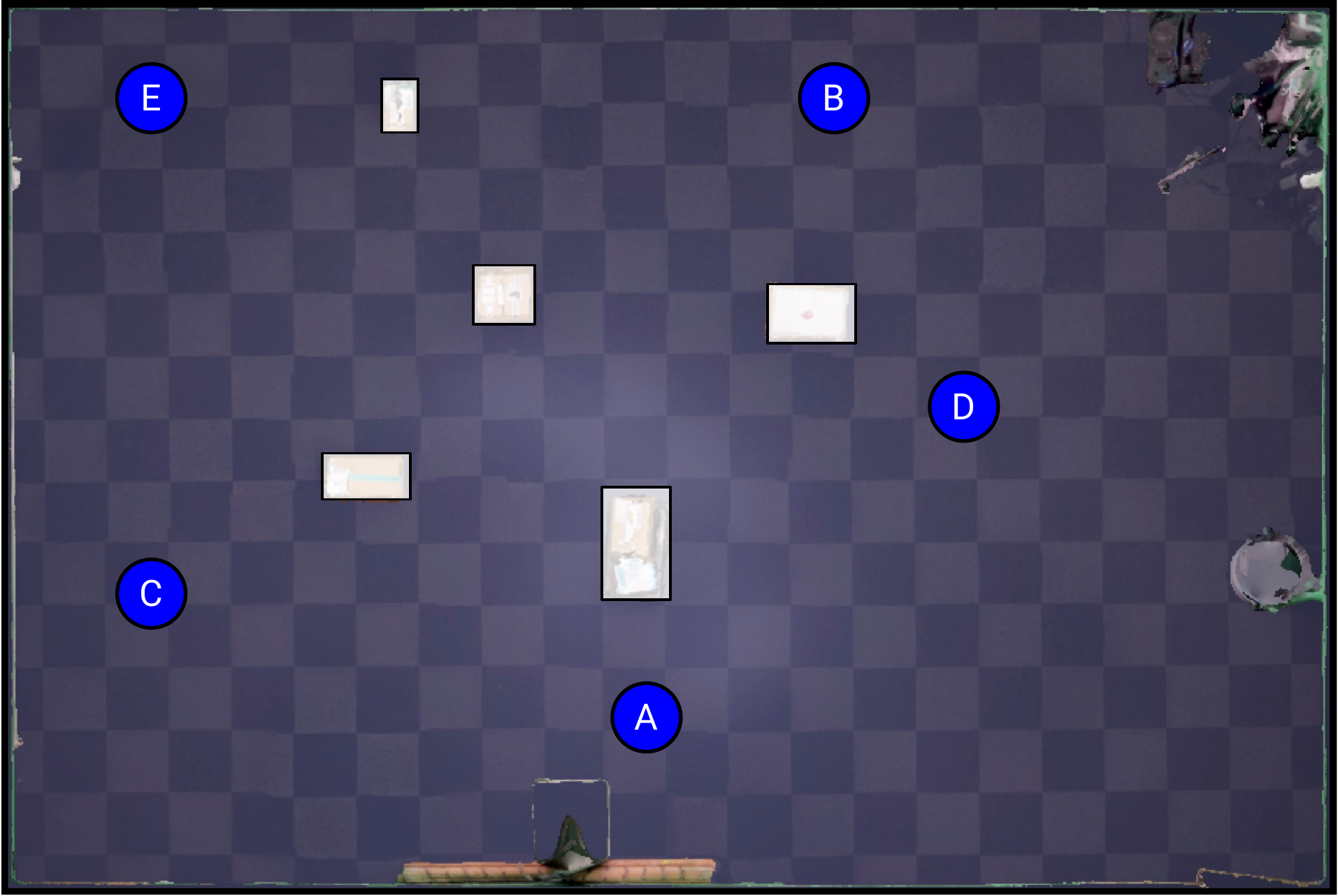}
    \caption{Top-down view of one of our testing environments. White boxes are obstacles. The robot navigates sequentially through the waypoints $A \rightarrow B \rightarrow C \rightarrow D \rightarrow E \rightarrow A$.}
    \label{fig:envconfigs}
\vspace{-.1cm}
\end{figure}

\vspace{-.1cm}
\subsection{Evaluation Environment}
\label{sec:evalenv}
Our evaluation environment is a controlled $6.5\text{m}$ by $10\text{m}$ interior room called CODA. Note that the agent was trained entirely in simulation and has never seen our evaluation environment room during training. The grid-like texture of the floor is purely coincidental, and not relied upon by the agent for navigation.
We create 3 room configurations (easy, medium, hard difficulty) with increasing number of tall, cardboard obstacles spread throughout the room. 
These `boxy' obstacles can be sensed easily by both the LIDAR and camera despite the two being vertically separated by \textasciitilde$0.5$m.
Objects like tables or chairs have narrow support at the LIDAR's height.
For each room configuration, we define a set of $5$ waypoints to serve as the start and end locations for navigation episodes.
\Cref{fig:envconfigs} shows top-down views of these room configurations. 

\xhdr{Virtualization.}
We digitize each environment configuration using a Matterport Pro2 3D camera to collect $360^\circ$ scans at multiple points in the room, ensuring full coverage. 
These scans are used to reconstruct 3D meshes of the environment which can be directly imported into Habitat.
This streamlined process is easily scalable and enables quick virtualization of new physical spaces.
On average, each configuration was reconstructed from $7$ panoramic captures and took approximately $10$ minutes.
We also evaluated how the reconstruction quality in simulation affects the transfer, this was done by dropping 5\% of mesh triangles in simulation
which led to a 23\% drop in SRCC (defined in Sec. IV.E).

\xhdr{Test protocol.}
We run parallel episodes 
in both simulation and reality. 
The agent navigates sequentially through the waypoints shown in \figref{fig:envconfigs} 
$(A \rightarrow B \rightarrow C \rightarrow D 
\rightarrow E \rightarrow A)$
for a total of 5 navigation episodes per room configuration. 
The starting points, starting rotations, and goal locations are identical across simulation and reality.
In total, we test 9 navigation models (described in the next section), 
in $3$ different room configurations, each with $5$ spawn-to-goal waypoints, and $3$ independent trials, 
for a total of $810$ runs in simulation and reality combined. Each spawn-to-goal navigation with \locobot 
takes approximately $6$ minutes, corresponding
to 40.5 hours of real-world testing. Safety guidelines 
require that a human monitor the experiments and at 8 hours a day, these experiments would take 5 days. 
With such 
long turn-around times, it is essential that we use
a robust pipeline to automate (or semi-automate) 
our experiments and 
reduce the cognitive load on the human supervisor. 
After each episode, the robot is automatically reset and has no knowledge from its previous run. 
For unsuccessful episodes, the robot uses a prebuilt environment map to navigate to the next episode start position. The room is equipped with a wireless camera 
to remotely track the experiments. 
In future, we plan to connect this automated 
evaluation setup to a docker2robot challenge, where participants can push 
code to a repository, which is then automatically evaluated on a real robot in this lab environment.

\subsection{Sim2Real Correlation Coefficient}
\label{sec:srcc}

To quantify the degree to which performance in simulation translates to performance in reality, we use a measure we call Sim2Real Correlation Coefficient (SRCC). 
Let $(s_i, r_i)$ denote 
accuracy (episode success rate, SPL \cite{anderson2018vision}, \etc) of navigation 
method $i$ in simulation and reality respectively. 
Given a paired dataset of accuracies for $n$ navigation methods $\{(s_1, r_1), \ldots, (s_n, r_n)\}$, SRCC is the sample 
Pearson correlation coefficient (bivariate correlation).\footnote{Other metrics such as rank correlation can also be used.}



SRCC values close to $+1$ indicate high linear correlation and are desirable, insofar as changes in simulation performance metrics correlate highly with changes in reality performance metrics. Values close to $0$ indicate low correlation and are undesirable as they indicate changes of performance in simulation is not predictive of real world changes in performance. Note that this definition of SRCC also suggests an intuitive visualization: by plotting performance in reality against performance in simulation as a scatterplot we can reveal the existence of performance correlation and detect outlier simulation settings or evaluation scenarios.

Beyond the utility of SRCC as a simulation predictivity metric, we can also view it as an optimization objective for simulation parameters. 
%
Concretely, let $\theta$ denote parameters controlling the simulator (amount of actuation noise, lighting, \etc).
We can view simulator design as optimization problem: 
$\max_\theta \text{SRCC}(S_n(\theta), R_n)$ where $S_n(\theta) = \{s_1(\theta), \ldots, s_n(\theta)\}$ is the set of accuracies in simulation with parameters $\theta$ and $R_n$ is the same performance metric computed on equivalent episodes in reality. 
Note that $\theta$ affects performance in simulation $S_n(\theta)$ but not $R_n$ since we are only changing test-time parameters. 
The specific navigation models themselves are held fixed. Overall, this gives us a formal approach to simulator design instead of operating on intuitions and qualitative assessments. 

In contrast, if a simulator has low SRCC but high mean real world performance, researchers will not be able to use this simulator to make decisions (e.g. model selection) because they can’t know if changes to performance in simulation will have a positive or negative effect on real-world performance. Every change will have to be tested on the physical robot. 


\section{Measuring the Sim2Real Gap}


\xhdr{Navigation Models.} 
We experiment with learning-based navigation models.  
Specifically, we train for PointGoal in Habitat on the $72$ Gibson environments that were rated $4+$ for quality in \cite{habitat19iccv}. 
For consistency with \cite{habitat19iccv}, we use the Gibson dataset%
 \cite{xia2018gibson} for training.
Agents are trained from scratch with reinforcement learning using DD-PPO~\cite{ddppo} -- a decentralized, distributed proximal policy optimization~\cite{schulman2017ppo} algorithm that is well-suited for GPU-intensive simulator-based training.
Each model is trained for $500$ million steps on $64$ Tesla V100s. For evaluation, we select the model with best Gibson-val performance.
We use the agent architecture from \cite{ddppo} 
composed of a visual encoder (ResNet50~\cite{he2016resnet}) and 
policy network (2-layer LSTM~\cite{hochreiter97lstm} with $512$-dim state).

\begin{figure}
\centering
\includegraphics[width=0.95\columnwidth]{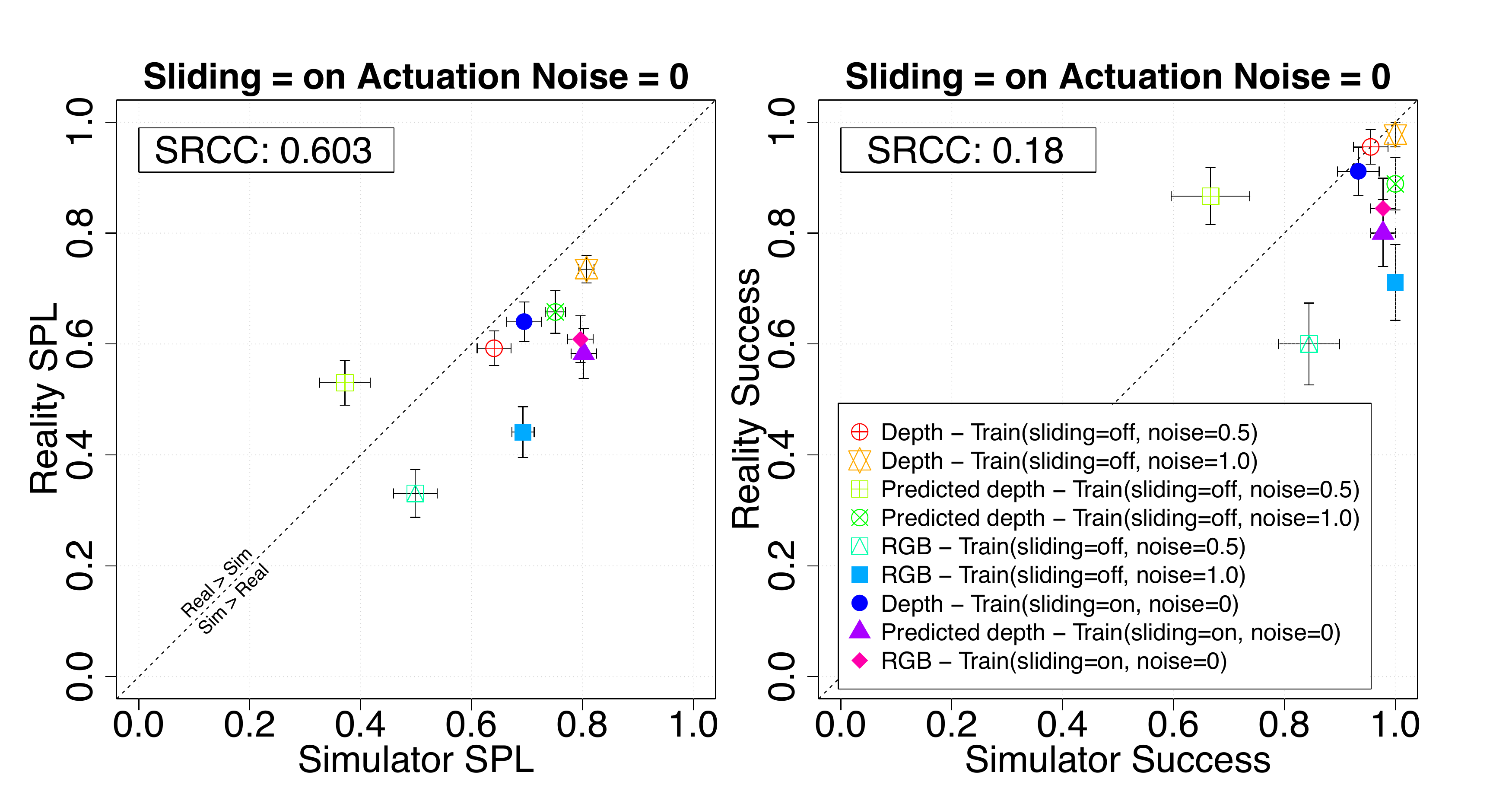}
\caption{\srccSPL (left) and \srccSucc (right) plots for AI Habitat Challenge 2019 test-sim setting in the CODA environment. We note a relatively low correlation between real and simulated performance.}
\label{fig:hab_scatter}
\end{figure}

Consistent with prior work, we train agents with RGB and depth sensors. 
Real-world depth sensors exhibit significant noise and are limited in range.
Thus, inspired by the winning entry in the RGB track of the Habitat Challenge 2019~\cite{habitat_challenge}, we also test an agent that uses a monocular depth estimator~\cite{icra_2019_fastdepth} to predict depth from RGB, which is then fed to the navigation model. 
In total, we train 9 different agents by varying sensor modalities (RGB, Depth, RGB$\rightarrow$Predicted Depth)  and training simulator configurations (\eg actuation noise levels). The exact settings 
are listed in Table \ref{tab:simvreal_results}. 
Note that the simulator parameters used for training these models may differ from the simulator parameters used for testing ($\theta$). 
Our goal is to span the spectrum of performance at test-time. 

\subsection{Revisiting the AI Habitat Challenge 2019}

\Cref{fig:hab_scatter} plots sim-vs-real performance of these 9 navigation models 
\wrt success rate (right) and SPL~\cite{anderson2018evaluation} (left).
Horizontal and vertical bars on a symbol indicate 
the standard error in simulation and reality respectively. 

\srccSPL is $0.60$, which is reasonably high but far from a level where we can be confident about evaluations in simulation alone. Problematically, there are 9 relative ordering reversals from simulation to reality. 
The success scatterplot (right) shows an even more disturbing trend -- nearly all methods (except one) appear to be working \emph{exceedingly} well in simulation with success rates close to 1.
However, there is a large dynamic range in success rates in reality.
This is summarized by a low \srccSucc of $0.18$, suggesting that improvements in performance in simulation are not predictive of performance improvements on a real robot.

Note that other than largely cosmetic adjustments to the robot size, sensor and action space specification, this simulation setting is not fundamentally different from the Habitat Challenge 2019. 
Upon deeper investigation, we discovered that one factor leading to this low sim2real predictivity is due to a `sliding' behavior in Habitat.


\begin{figure}
    \centering
    \includegraphics[width=0.95\columnwidth]{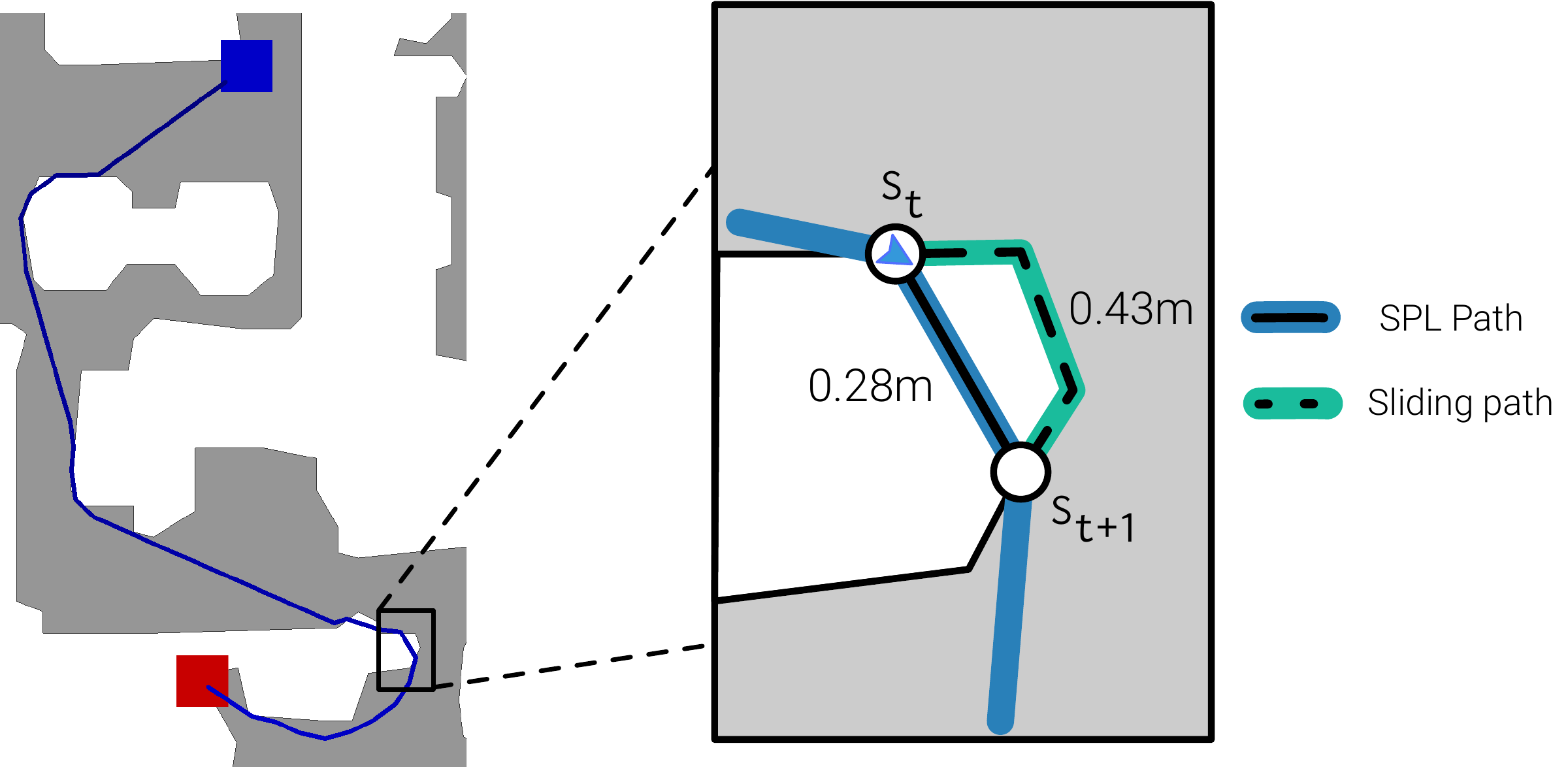}
    \caption{Sliding behavior leading to `cheating' agents. At time $t$, the agent at $s_t$ executes a \texttt{forward} action, and slides along the wall to state $s_{t+1}$. The resulting straight-line path (used to calculate SPL) goes outside the environment. Gray denotes navigable space while white is non-navigable.}
    \label{fig:corner-clipping}
\end{figure}

\xhdr{Cheating by sliding.}
In Habitat-Sim~\cite{habitat19iccv}, when the agent takes an action that results in a collision, the agent \emph{slides} along the obstacle as opposed to stopping. This behavior is also seen in many other simulators -- it is enabled by default in MINOS~\cite{savva2017minos}, and Deepmind Lab~\cite{beattie2016deepmind}, and is also prevalent in simulators and video game engines that employ physics backend, such as Gibson \cite{xia2018gibson}, and AI2 THOR~\cite{ai2thor}. This is because there is no perfect physics simulator, only approximations; this type of physics allows for smooth human control, but does not accurately reflect real-world physics nor safety precautions the robot platform may employ (\ie stopping on collision). 
We find that this enables `cheating' by learned agents.
As illustrated by an example in \figref{fig:corner-clipping}, 
the agent exploits this sliding mechanism to take an effective path that appears to travel \emph{through non-navigable regions} of the environment (like walls). Let $s_t$ denote agent position at time $t$, where the agent is already in contact with an obstacle.
The agent executes a \texttt{forward} action, collides, and slides along the obstacle to state $s_{t+1}$.  
The path taken during this maneuver is far from a straight line, however for the purposes of computing 
SPL (the metric the agent optimizes), 
Habitat calculates the Euclidean distance travelled 
$||s_t - s_{t+1}||_2$.
This is equivalent to taking a straight line path between $s_t$ and $s_{t+1}$ that goes outside the navigable regions of the environment 
(appearing to go through obstacles). 
On the one hand, the emergence of such exploits is a 
sign of success of large-scale reinforcement learning -- clearly, we are maximizing reward. On the other hand, this is a problem for sim2real transfer. 
Such policies fail disastrously in the real world where the robot bump sensors force a stop on contact with obstacles.
To rectify this issue, we modify Habitat-Sim to 
disable sliding on collisions. 
The discovery of this issue motivated our investigation into optimizing simulation parameters. 




\begin{figure}
\includegraphics[width=\columnwidth]{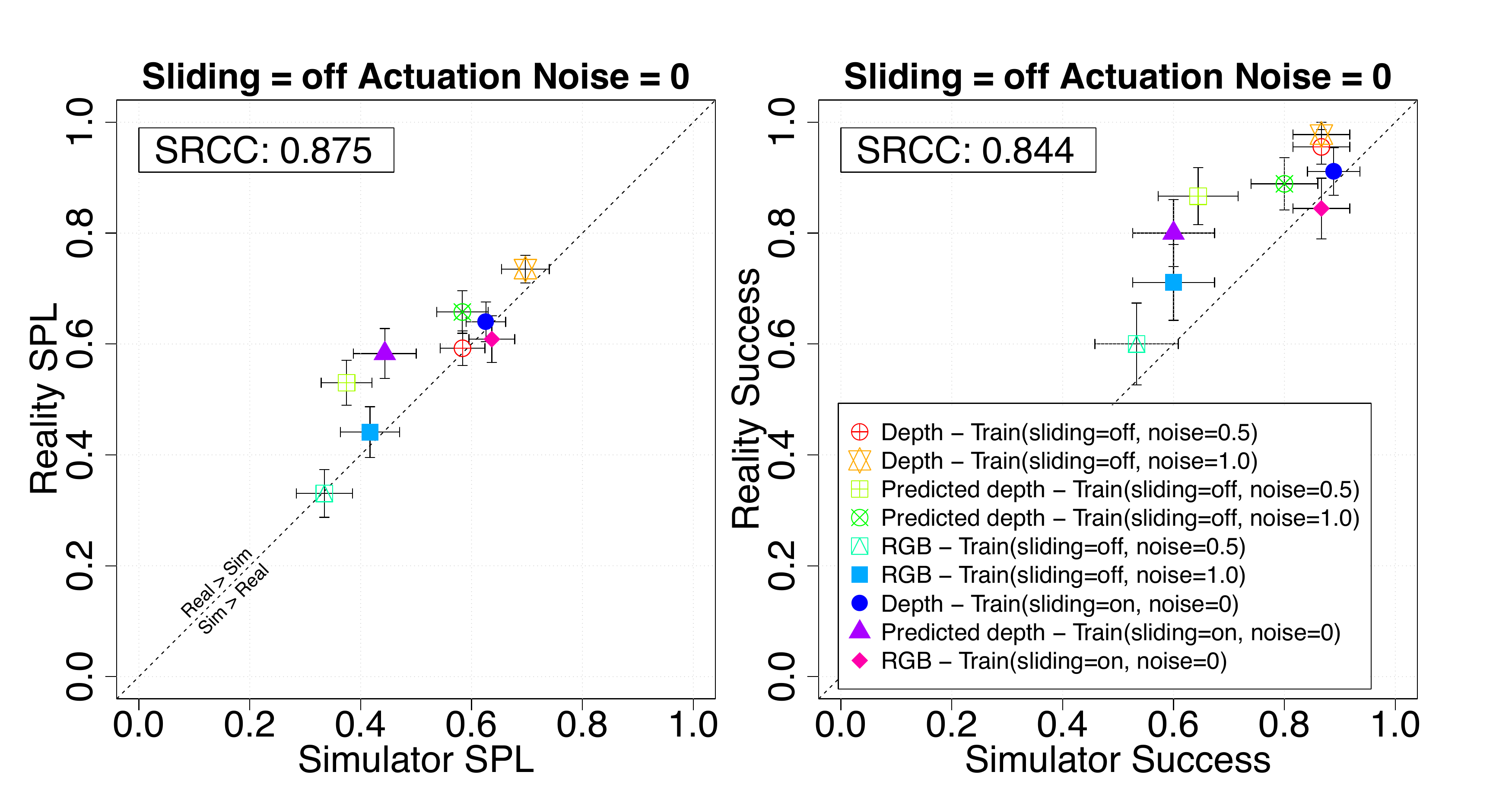}
\caption{Optimized \srccSPL (left) and \srccSucc (right) scatterplots in the CODA environment. Comparing with \Cref{fig:hab_scatter} we see improvements, indicating better predictivity of real-world performance.}
\label{fig:scatteroptimized}
\end{figure}

\subsection{Optimizing simulation with SRCC}

We perform grid-search over simulation parameters -- 
sliding (off vs on) and a scalar multiplier on actuation noise (varying from 0 to 1, in increments of 0.1). 
We find that sliding off and $0.0$ actuation noise lead to the highest SRCC. 
\figref{fig:scatteroptimized} shows 
a remarkable alignment in the SPL scatter-plot (left) -- nearly all models lie close to the diagonal, 
suggesting that we fairly accurately predict how a model is going to perform on the robot by testing in simulation. 
Recall that the former takes $40.5$ hours and significant human effort, while the latter is computed in under 1 minute. 
\srccSPL improves from $0.603$ (\figref{fig:hab_scatter}) to $0.875$ ($0.272$ gain). 
\srccSucc shows an even stronger improvement -- 
from $0.18$ to $0.844$ ($0.664$ gain)! 

\begin{table}
\vspace{5pt}
  \renewcommand\theadalign{bc}%
  \renewcommand\theadfont{\bfseries}%
  \renewcommand\theadgape{\Gape[2pt]}%
  \renewcommand\cellgape{\Gape[2pt]}%
  \renewcommand{\tableTitle}[1]{\cellcolor{white}{\parbox[t]{2.5mm}{\multirow{-9}{*}{\rotatebox[origin=c]{90}{\cellcolor{white}\textbf{#1}}}}}}%
  \setlength{\tabcolsep}{3pt}%

\caption{Average performance in reality (col.~3), CODA and Gibson scenes under Habitat Challenge 2019 settings [sliding=on, noise=0.0] (col.~5, col.~7), and test-sim [sliding=off, noise=0.0] (col.~6, col.~8) across different train-sim configurations (col.~1-2).}
\label{tab:simvreal_results}
  \scriptsize
  \centering%
   \rowcolors{2}{white}{gray!25}%
    \begin{tabularx}{\linewidth}{cYYYYYYY}
      \toprule
      \rowcolor{white} \textbf{Sensor}  & \textbf{Train-Sim Noise} & \textbf{Train-Sim Sliding} & \textbf{Reality SPL} & \textbf{CODA Chall-Sim SPL} & \textbf{CODA Test-Sim SPL} &\textbf{Gibson Chall-Sim  SPL} & \textbf{Gibson Test-Sim SPL}\\
      \midrule
       Depth & 0.5 & off & 0.59 & 0.64 & 0.58 & 0.68 & 0.59\\
       Depth & 1.0 & off & 0.74 & 0.81 & 0.70 & 0.78 & 0.53\\
       Pred. Depth & 0.5 & off & 0.53 & 0.37 & 0.37 & 0.54 & 0.40\\
       Pred. Depth & 1.0 & off & 0.66 & 0.75 & 0.58 & 0.64 & 0.43\\
       RGB & 0.5 & off & 0.33 & 0.50 & 0.33 & 0.56 & 0.43\\
       RGB & 1.0 & off & 0.44 & 0.69 & 0.42 & 0.58 & 0.36\\
       Depth & 0.0 & on & 0.64 & 0.70 & 0.63 & 0.66 & 0.35\\
       Pred. Depth & 0.0 & on & 0.58 & 0.80 & 0.44 & 0.56 & 0.32\\
       RGB & 0.0 & on & 0.61 & 0.80 & 0.64 & 0.62 & 0.36\\\bottomrule
    \end{tabularx}
\end{table}

From \Cref{tab:simvreal_results}, we can see that the best performance in reality (Reality SPL) is achieved by row 2, which is confidently predicted by CODA Test-Sim SPL (col 4) by a strong margin. 
The fact that no actuation noise is the optimal setting suggests that our chosen actuation noise model (from PyRobot~\cite{pyrobot2019}) may not reflect conditions in reality.

To demonstrate the capabilities of our models in navigating cluttered environments,  
we also evaluate performance 
on the Gibson dataset,
which contains 
$572$ scanned spaces (apartments, multi-level homes, offices, houses, hospitals, gyms), 
containing furniture (chairs, desks, sofas, tables),  
for a total of $1447$ floors. 
We use the `Gibson-val' navigation episodes from Savva \emph{et al.}~[1]. 
We observe a similar drop between performance  
between sliding on (Chall-Sim col 5, 7 in Table I) vs sliding off (Test-Sim col 4,6 in Table I). 
This further suggests that the participants in the Habitat Challenge 2019 would not achieve similar performance on a real robot.

We also implement a wall-following oracle, 
and compare its performance to our learned model (Depth, actuation noise=0.5, sliding=off). 
The oracle receives full visibility of the environment via 
a top-down map of the environment. Note that this is significantly stronger input 
than our learned models (which operate on egocentric depth frames). 
The oracle navigates to the goal by following a 
straight path towards the goal and follows walls upon coming in contact with obstacles. 
We endow the oracle with perfect wall-following (\emph{i.e.}   
it follows the wall along the shortest direction to the goal), 
and it never actually collides or gets stuck on obstacles. 
This oracle achieves 0.46 SPL on Gibson-val, compared to 0.59 SPL
achieved by our model under the Gibson Test-Sim setting. 
For episodes longer 
than 10m, the oracle performance drops to 0.05 SPL, compared to our model, 
which achieves a 0.23 SPL. 
These experiments show the need for a learning based approach in complex and 
cluttered environments.

\section{Conclusion}


We introduce the Habitat-PyRobot Bridge (HaPy) library that allows for seamless deployment of visual navigation models.
Using Matterport scanning, the Habitat stack, the HaPy library, and \locobot we benchmark the correlation between reality and simulation performance with the SRCC metric.
We find that naive simulation parameters lead to low correlation between performance in simulation and performance in reality.
We then optimize the simulation parameters for SRCC and obtain high predictivity of real world performance without running new evaluations in the real world.
We hope that the infrastructure we introduce and the conceptual framework of optimizing simulation for predictivity will enable sim2real transfer in a variety of navigation tasks.

\xhdr{Acknowledgements.} We thank the reviewers for their helpful suggestions. We are grateful to Kalyan Alwala, Dhiraj Gandhi and Wojciech Galuba for their help and support. The Georgia Tech effort was supported in part by NSF, AFRL, DARPA, ONR YIPs, ARO PECASE, Amazon. The views and conclusions contained herein are those of the authors and should not be interpreted as necessarily representing the official policies or endorsements, either expressed or implied, of the U.S. Government, or any sponsor.
\\
\xhdr{Licenses for referenced datasets.} \\
{\footnotesize
Gibson: \url{https://storage.googleapis.com/gibson_material/Agreement\%20GDS\%2006-04-18.pdf}
\\
Matterport3D: \url{http://kaldir.vc.in.tum.de/matterport/MP_TOS.pdf}
}

\vspace{-0.08in}
\LetLtxMacro{\section}{\oldsection}
\vspace{-0.06in}
{
\bibliographystyle{style/IEEEtran}
\bibliography{bib/strings,bib/main}
}
\end{document}